\documentclass[10pt,twocolumn,letterpaper]{article}

\usepackage{iccv}
\usepackage{times}
\usepackage{epsfig}
\usepackage{graphicx}
\usepackage{amsmath}
\usepackage{amssymb}

\usepackage{graphicx}
\usepackage{amsmath}
\usepackage{amssymb}
\usepackage{booktabs}
\usepackage{romannum} 
\usepackage{algorithm} 
\usepackage{algpseudocode} 
\usepackage{wrapfig}
\usepackage{arydshln} 
\usepackage{makecell}
\usepackage{multirow}
\usepackage{pifont}
\def\etal{\emph{et al.~}} 
\def\eg{\emph{e.g}\onedot{\emph{,~}}}  
\def\ie{\emph{i.e}\onedot{\emph{,~}}}  

\usepackage{amsmath,amsfonts,bm}

\def\eqref#1{equation~\ref{#1}}

\def\1{\bm{1}}

\DeclareMathAlphabet{\mathsfit}{\encodingdefault}{\sfdefault}{m}{sl}
\SetMathAlphabet{\mathsfit}{bold}{\encodingdefault}{\sfdefault}{bx}{n}

\newcommand{\E}{\mathbb{E}}

\usepackage{dsfont}
\usepackage{arydshln}
\usepackage{enumitem}
\usepackage[breaklinks=true,bookmarks=false]{hyperref}

\iccvfinalcopy 

\def\iccvPaperID{****} 

\ificcvfinal\fi

\begin{document}

\title{Mitigating Adversarial Vulnerability through Causal Parameter Estimation\\by Adversarial Double Machine Learning}


\author{Byung-Kwan Lee\thanks{Equal contribution. $\dagger$ Corresponding author.},~~Junho Kim\footnote[1]{},~~Yong Man Ro\footnote[2]{}\\
Image and Video Systems Lab, School of Electrical Engineering, KAIST, South Korea\\
{\tt\small \{leebk, arkimjh, ymro\}@kaist.ac.kr}
}

\maketitle
\ificcvfinal\thispagestyle{empty}\fi

\begin{abstract}
   Adversarial examples derived from deliberately crafted perturbations on visual inputs can easily harm decision process of deep neural networks. To prevent potential threats, various adversarial training-based defense methods have grown rapidly and become a de facto standard approach for robustness. Despite recent competitive achievements, we observe that adversarial vulnerability varies across targets and certain vulnerabilities remain prevalent. Intriguingly, such peculiar phenomenon cannot be relieved even with deeper architectures and advanced defense methods. To address this issue, in this paper, we introduce a causal approach called Adversarial Double Machine Learning (ADML), which allows us to quantify the degree of adversarial vulnerability for network predictions and capture the effect of treatments on outcome of interests. ADML can directly estimate causal parameter of adversarial perturbations per se and mitigate negative effects that can potentially damage robustness, bridging a causal perspective into the adversarial vulnerability. Through extensive experiments on various CNN and Transformer architectures, we corroborate that ADML improves adversarial robustness with large margins and relieve the empirical observation.
\end{abstract}

\section{Introduction}
\label{sec:intro}
Along with the progressive developments of deep neural networks (DNNs)~\cite{resnet, dosovitskiy2020image, carion2020end}, an aspect of AI safety comes into a prominence in various computer vision research~\cite{samangouei2018defense, zhang2018camou, eykholt2018robust, jagielski2018manipulating}. Especially, adversarial examples~\cite{42503, 43405, lee2022masking} are known as potential threats on AI systems. With deliberately crafted perturbations on the visual inputs, adversarial examples are hardly distinguishable to human observers, but they easily result in misleading decision process of DNNs. Such adversarial vulnerability provokes weak reliability of inference process of DNNs and discourages AI adoption to the safety critical areas~\cite{WANG201912, 8824956}.

\begin{figure}[t!]
\centering
\includegraphics[width=0.99\linewidth]{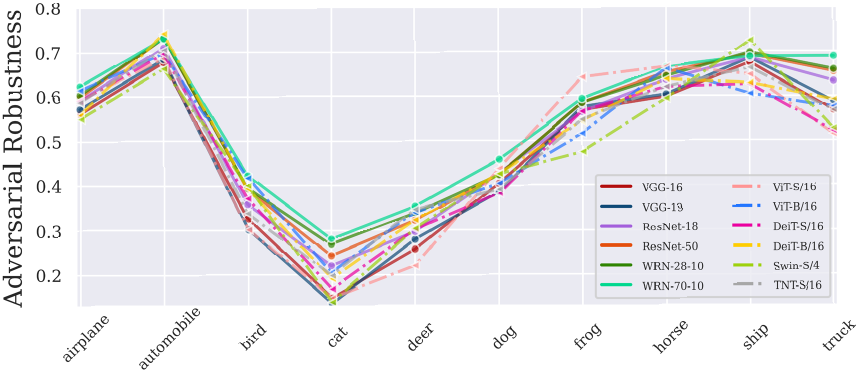}
\vspace*{-0.55cm}
\begin{flushleft}
    \hspace{2.5cm}{(a) Network Architectures}
\end{flushleft}	
\vspace*{-0.35cm}
\includegraphics[width=0.94\linewidth]{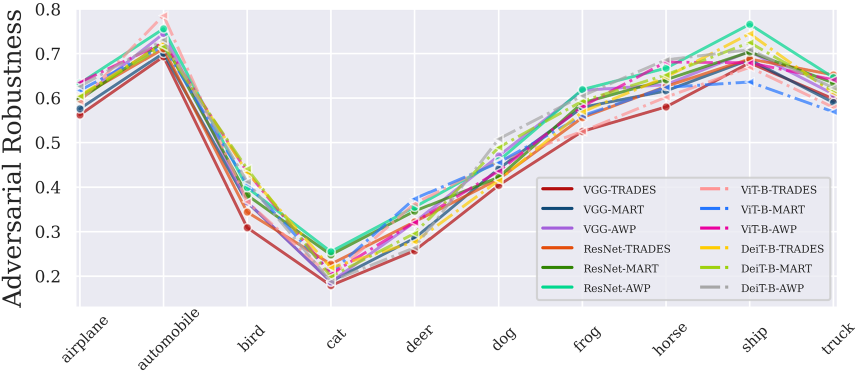}
\vspace*{-0.5cm}
\begin{flushleft}
    \hspace{1.9cm}{(b) Adversarial Defense Methods}
\end{flushleft}	
\vspace*{-0.25cm}
\caption{The comparison of adversarial robustness along target classes with respect to (a) Network Architectures and (b) Adversarial Defense Methods on CIFAR-10~\cite{krizhevsky2009learning}. Note that, the distribution of adversarial robustness is consistent along both criteria.}
\label{fig:intro}
\vspace{-0.6cm}
\end{figure}

In order to achieve robust and trustworthy DNNs from adversarial perturbation, previous methods~\cite{madry2018towards, 45816, CW, pmlr-v97-zhang19p, Wang2020Improving, wu2020adversarial, croce2020reliable} have delved into developing various adversarial attack and defense algorithms in the sense of cat-and-mouse game. As a seminal work, Madry~\etal~\cite{madry2018towards} have paved the way for obtaining robust network through adversarial training (AT) regarded as an ultimate augmentation training~\cite{tsipras2018robustness} with respect to adversarial examples. Based on its effectiveness, various subsequent works~\cite{pmlr-v97-zhang19p, Wang2020Improving, wu2020adversarial, zhang2021geometryaware, rade2022reducing, kim2023demystifying} have investigated it to further enhance adversarial robustness. 

Although several AT-based defense methods have become a de facto standard due to their competitive adversarial robustness, we found an intriguing property of the current defense methods. As in Figure~\ref{fig:intro}, we identify that the adversarial robustness for the each target class significantly varies with a large gap, and this phenomenon equally happens in the course of (a) network architectures and (b) various AT-based defense methods. In addition, we would like to point out that the robustness of particular target is still severely vulnerable than others even with advanced architectures~\cite{dosovitskiy2020image, touvron2021training, liu2021swin} and defense methods~\cite{pmlr-v97-zhang19p, Wang2020Improving, wu2020adversarial}. We argue that such phenomenon is derived from the current learning strategies of AT-based defense methods that lacks of understanding causal relations between the visual inputs and predictions. When considering AT methods as the ultimate augmentation~\cite{tsipras2018robustness}, current methods rely solely on strengthening the correlation between adversarial examples and target classes through canonical objectives that improve robustness. To fundamentally address such vulnerability and understand the causal relation, we need to quantify the degree of vulnerability (\ie causal parameter) and should mitigate its direct effects to the network predictions.

\begin{figure}[t!]
\centering
\includegraphics[width=0.99\linewidth]{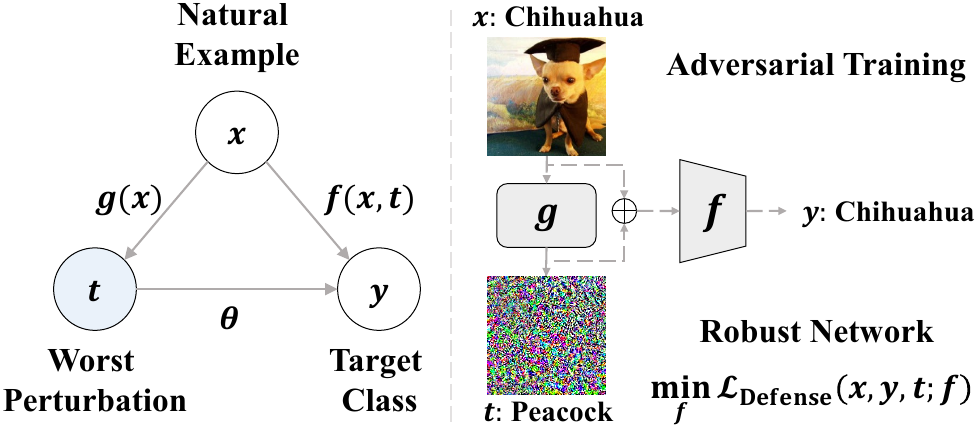}
\vspace*{-0.3cm}
\begin{flushleft}
    \hspace{0.18cm}{(a) AT Causal Diagram \hspace{0.5cm}(b) Training Robust Network}
\end{flushleft}	
\vspace*{-0.2cm}
\caption{Overview of canonical adversarial training procedure for robust network and its causal diagram.}
\label{fig:dgp}
\vspace{-0.6cm}
\end{figure}

Accordingly, we investigate the AT-based defense methods in a causal viewpoint and propose a way of precisely estimating causal parameter between adversarial examples and their predictions, namely \textit{Adversarial Double Machine Learning (ADML)}. We first represent a causal diagram of AT-based methods and interpret it as a generating process of robust classifiers as illustrated in Figure~\ref{fig:dgp}. Regarding standard adversarial training~\cite{madry2018towards} as an optimizing procedure for the robust network parameters $f$ with respect to the worst perturbations $t$, we can instantiate a generation $g$\footnote{Selecting $g$ as proper perturbations varies according to domain specific tasks (\eg rotations, translations~\cite{engstrom2018rotation}, or spatial deformations~\cite{xiao2018spatially}).} as an adversarial attack of projected gradient descent (PGD)~\cite{madry2018towards} for the given clean examples $x$. 

Then, our research question is how to quantitatively compute the causal parameter $\theta$ between the perturbations $t$ and target classes $y$, and identify the causal effects on outcome of our interests. Through double machine learning (DML)~\cite{chernozhukov2018double}, widely studied as a powerful causal estimator~\cite{chernozhukov2017double, colangelo2020double, pmlr-v162-fingerhut22a, jung2021double, jung2021estimating} for the given two regression models, we can establish an initial research point of estimating causal parameter of adversarial perturbation with theoretical background. However, it is difficult to directly estimate $\theta$ in the high-dimensional manifolds, especially for DNNs. In this paper, we shed some lights on identifying causal parameter of the perturbations while theoretically bridging the gap between causal inference and adversarial robustness. Then, by minimizing the magnitude of the estimated causal parameter, we essentially lessen negative causal effects of adversarial vulnerability, and consequently acquire robust network with the aforementioned phenomenon alleviated.

To corroborate the effectiveness of ADML on adversarial robustness, we set extensive experiments with four publicly available datasets~\cite{krizhevsky2009learning, le2015tiny, deng2009imagenet}. Our experiments include various convolutional neural network architectures (CNNs), as well as Transformer architectures that have drawn great attention in both vision and language tasks~\cite{vaswani2017attention, dosovitskiy2020image, zhu2020deformable, zhao2021point} yet relatively lack of being studied in adversarial research.

Our contributions can be summarized as follows:
\begin{itemize}[nosep]
\item We present an empirical evidence that despite the recent advances in AT-based defenses, fundamentally adversarial vulnerability still remains across various architectures and defense algorithms.
\item Bridging a causal perspective into adversary, we propose Adversarial Double Machine Learning (ADML), estimating causal parameter in adversarial examples and mitigating its causal effects damaging robustness.
\item Through extensive experiments and analyses on various CNN and Transformer architectures, we corroborate intensive robustness of our proposed method with the phenomenon alleviated. 
\end{itemize}


\section{Background and Related Work}
\label{sec:related}

\noindent\textbf{Notation.~} We deal with DNNs for classification as in Figure~\ref{fig:dgp}, represented by $f:\mathcal{X} \rightarrow \mathcal{Y}$, where $\mathcal{X}$ and $\mathcal{Y}$ denotes image and probability space, respectively. Let $x\in \mathcal{X}$ denote clean images and $y\in\mathcal{Y}$ indicate (one-hot) target classes corresponding to the images. Adversarial examples $x_{\text{adv}}$ are generated by adversarial perturbations $t$ through DNNs, such that $x_{\text{adv}}=x+t$. Here, the perturbations are carefully created through the following formulation:
\begin{equation}
\label{eqn:attack}
\max\limits_{\left \| t \right \|_{\infty} \leq \gamma}\mathcal{L}_\text{CE}(f(x+t),y),
\end{equation}
where $\mathcal{L}_\text{CE}$ represents a pre-defined loss such as cross-entropy for classification task. We regard adversarial perturbations $t$ as $l_{\infty}$ perturbation within $\gamma$-ball (\ie perturbation budget). Here, $\left \| \cdot \right \|_{\infty}$ describes $l_{\infty}$ perturbation magnitude.

\subsection{Adversarial Training}

After several works~\cite{42503, 43405,kim2021distilling} have found that human-imperceptible adversarial examples easily break network predictions, Madry~\etal~\cite{madry2018towards} have thrown a fundamental question: ``How can we make models robust to adversarial examples with security guarantee?''. To answer it, they have introduced the concept of empirical risk minimization (ERM) serving as a recipe to obtain classifiers with small population risk. Thanks to its reliable guarantee, they have consolidated it on the purpose of adversarial defense and accomplished the yardstick of adversarial training. The key factor of its achievement is regarding adversarial training as min-max optimization in a perspective of saddle point problem, which can be written as follows: 
\begin{equation}
\label{eqn:defense}
    \min\limits_{f} \E_{(x, y)\sim\mathcal{D}}\left[ \max\limits_{\left \| t \right \|_{\infty} \leq \gamma}  \mathcal{L}_{\text{CE}} \left(f(x+t),y\right) \right],
\end{equation}
where $\mathcal{D}$ denotes a set of data samples $(x, y)$. Here, they have presented an adversarial attack based on PGD to powerfully behave inner-maximization on Eq.~(\ref{eqn:defense}), which is an ultimate first-order adversary with a multi-step variant of fast gradient sign method~\cite{45816} by adding a random perturbation around the clean images $x$.

According to its impact, various adversarial training methods~\cite{madry2018towards, pmlr-v97-zhang19p, Wang2020Improving, wu2020adversarial} have grown exponentially and become de facto standards robustifying DNNs against adversarial perturbation. Zhang~\etal~\cite{pmlr-v97-zhang19p} have pointed out the trade-off between clean accuracy and adversarial robustness, and reduced the gap between clean errors and robust errors. Wang~\etal~\cite{Wang2020Improving} have claimed that all of clean images are used to perform both inner-maximization and outer-minimization process in Eq.~(\ref{eqn:defense}), irrespective of whether they are correctly classified or not. Thus, they have focused on misclassified clean images prone to be easily overlooked during adversarial training and demonstrated their significant impacts on the robustness by incorporating an explicit regularizer for them. Wu~\etal~\cite{wu2020adversarial} have studied loss landscapes with respect to network parameters and shown a positive correlation between the flatness of the parameter loss landscapes and the robustness. In the end, they have presented a double-perturbation mechanism where clean images are perturbed, while network parameters are simultaneously perturbed as well.

On the other hand, we plunge into investigating where adversarial vulnerability comes from and observe that the vulnerability varies along target classes, and it significantly deteriorates network predictions. Further, we find that this phenomenon commonly happens across various network architectures and advanced defense methods. To relieve such peculiarity, we deploy double machine learning (DML) that helps to capture how treatments (\ie adversarial perturbations) affect outcomes of our interests (\ie network predictions), which is one of the powerful causal inference methods. After we concisely explicate the necessary background of DML, we will bridge it to the adversary in Sec.~(\ref{sec:proposed}).

\subsection{Double Machine Learning}
In data science and econometrics, one of the fundamental problems is how to measure causality between treatments $t$ and outcomes of our interest $y$ among high-dimensional observational data samples (see Figure~\ref{fig:dgp}) to identify data generating process. At a first glance, it seems simple to compute their causality, but we should keep in mind the possibility for the existence of covariates $x$ affecting both treatments and outcomes. In other words, for example, if one may want to know the causal effects of drug dosage $t$ to blood pressure changes $y$, one needs to collect observational data with respect to a variety of patients characteristics and their clinical histories $x$, so as not to fall into biased environment. In reality, though, it is impossible to collect observational data including all covariates concerning treatments and outcomes, so it is not an easy problem to catch genuine causality under the unknown covariates $x$. Therefore, there has been a growing demand for robustly predicting the unbiased causal relation, despite with the limited data samples.

Recently, the advent of double machine learning (DML)~\cite{chernozhukov2018double} enables us to clarify the causality between treatments $t$ and outcomes $y$, when two regression models are given. The formulation of initial DML can be written as:
\begin{equation}
\begin{split}
\label{eqn:partially}
       y &= f(x) + \theta t + u, \quad (\E[u\mid x, t]=0)\\
       t &= g(x) + v, \quad\quad\quad (\E[v \mid x]=0)
\end{split}
\end{equation}
where $\theta\in\mathbb{R}$ denotes causal parameter representing causal relation between $t\in\mathbb{R}^{d}$ and $y\in\mathbb{R}^{d}$. In addition, $f$ indicates one regression model projecting covariates to outcome domain, and $g$ denotes another regression model generating treatments $t$. In the sense that two regression models $f$ and $g$ are not main interest of DML, they are called as nuisance parameters to estimate the causal parameter $\theta$. Note that, early DML assumes the problem setup is proceeded in partially linear settings as a shape of Robinson-style~\cite{robinson1988root} described in Eq.~(\ref{eqn:partially}), where ``partially'' literally means that treatments $t\in\mathbb{R}^{d}$ are linearly connected to outcome $y\in\mathbb{R}^{d}$, while covariates $x$ are not. In addition, it is supposed that the conditional expected error of $u\in\mathbb{R}^{d}$ and $v\in\mathbb{R}^{d}$ equals to zero vector $0\in\mathbb{R}^{d}$.

To obtain the causal parameter $\theta$, Chernozhukov~\etal~\cite{chernozhukov2018double} have provided a solution of estimating the causal parameter such that $\hat{\theta} = (y-\E[y\mid x]) \cdot v / \|v\|^2$ which satisfies Neyman-orthogonality~\cite{neyman1965asymptotically, neyman1979c}. It makes $\hat{\theta}$ invariant to their erroneous of two nuisance parameters with the variance of causal parameter reduced. Furthermore, they have addressed a chronic problem that $\theta$ is only accessible when the two nuisance parameters are in a class of Donsker condition, where deep neural networks are not included in that condition. They have theoretically demonstrated \textit{sample-splitting} plus \textit{cross-fitting} can effectively relax Donsker condition and allow a broad array of modern ML methods~\cite{chernozhukov2018double} to compute unbiased causal parameter $\theta$. 


Following the principle, they first split the data samples $\{\mathcal{D}_1, \mathcal{D}_2\}\sim\mathcal{D}$ and divided the process of causal inference into two steps: (a) training two nuisance parameters $f$ and $g$ with $\mathcal{D}_1$, (b) estimating unbiased $\theta$ with $\mathcal{D}_2$. Here, data samples $\mathcal{D}_2$ used to estimate unbiased causal parameters should not be overlapped with $\mathcal{D}_1$ utilized to train the nuisance parameters. To make copious combinations, they swapped the role of partitioned data samples $\mathcal{D}_1\leftrightharpoons\mathcal{D}_2$ or repeatedly split $\mathcal{D}$. Subsequently, they have performed cross-fitting (\eg $k$-fold cross validation) by averaging the estimated causal parameters from various split samples. 

Along with the success of initial DML in partially linear settings, numerous variants~\cite{chernozhukov2017double, colangelo2020double, mackey2018orthogonal, foster2019orthogonal, pmlr-v162-fingerhut22a, klosin2021automatic, chernozhukov2022riesznet} have emerged, and they have extended its initial nature to non-parametric settings with continuous treatments $t$ in order to capture more complicated non-linear causal relations in a debiased state. A non-parametric formulation~\cite{chernozhukov2017double} represents a more general problem setup of DML as follows:
\begin{equation}
\begin{split}
\label{eqn:np_dml}
       y &= f(x, t) + u, \quad (\E[u\mid x, t]=0)\\
       t &= g(x) + v, \quad\quad (\E[v \mid x]=0)
\end{split}
\end{equation}
where there is no explicit term for causal parameter $\theta$ exhibiting causal relation between treatments $t$ and outcomes $y$, compared to Eq.~(\ref{eqn:partially}). Colangelo~\etal~\cite{colangelo2020double} have introduced a way of estimating causal parameter $\theta$ applicable to non-parametric settings with high-dimensional continuous treatments $t\in\mathcal{T}$, which can be written as:
\begin{equation}
\label{eqn:nl_theta}
\begin{split}
       \hat{\theta} = \frac{\partial}{\partial t}\E[y\mid \text{do}(\mathcal{T}{=}t)].
\end{split}
\end{equation}
They have utilized do-operator~\cite{pearl2009causality} commonly used in graphical causal models and intervened on treatments $t$ to compute an interventional expectation $\E[y|\text{do}(\mathcal{T}{=}t)]$. It represents the expected outcome averaged from all the possible covariates for the given fixed treatments $t$, such that $\E[y|\text{do}(\mathcal{T}{=}t)]=\sum_{x\in\mathcal{X}}\E[y| x,t]p(x)$. Specifically, they have estimated causal parameter $\theta$ by measuring how much the interventional expectation shifted, once they change the treatments slightly. Since the most important property of DML is Neyman-Orthogonality helping to robustly estimate the causal parameter, the interventional expectation should be also modified to satisfy the property~\cite{colangelo2020double, kennedy2020optimal} of its invariance to nuisance parameters $f$ and $g$. Its formulation can be written as follows (see details in Appendix A):
\begin{equation}
\begin{split}
\label{eqn:neyman_e_do_t}
       \E[y\mid \text{do}(\mathcal{T}{=}t)]=\E_{\mathcal{D}_{t}}\left[f(x,t)+\frac{y-f(x,t)}{p(\mathcal{T}{=}t\mid x)}\right],
\end{split}
\end{equation}
where $\mathcal{D}_t$ denotes a set of observational covariates and outcome samples for a fixed $t\in\mathcal{T}$ such that $(x,y)\sim\mathcal{D}_{t}$, a sub-population of $\mathcal{D}$. Note that, $p(\mathcal{T}{=}t| x)$ is related to treatment generator $g$.  Here, differentiating Eq.~(\ref{eqn:neyman_e_do_t}) enables us to acquire unbiased causal parameter in non-parametric settings with non-linear causal relation.

In brief, DML captures unbiased causal relation between treatments $t$ and outcomes $y$ even with finite data samples, of which theoretical ground is (a) Neyman-Orthogonality for robustly estimated causal parameter despite undesirable outputs of nuisance parameters, and (b) sample-splitting plus cross-fitting for debiased causal parameters.


\section{Adversarial Double Machine Learning}
\label{sec:proposed}


\subsection{Adversarial Data Generating Process}

In general deep learning schemes, we have clean visual images $x\in \mathbb{R}^{hwc}$ and their corresponding target classes $y\in\mathbb{R}^{d}$ in our hand as a format of dataset, where $h$, $w$, $c$ denotes image resolution of height, width, channel, repectively, and $d$ denotes the number of classes. Thus, we do not need additional data generating process. For adversarial training, on the other hand, we need another data, which are adversarial perturbations generated from data samples $(x,y)$ as in Eq.~(\ref{eqn:attack}). They are normally created by PGD~\cite{madry2018towards} at every training iteration to make DNNs $f$ robust through min-max optimization game according to Eq.~(\ref{eqn:defense}).

Though, the more iterations of adversarial training, the fewer perturbations that impair network predictions. In other words, not all of the perturbations can corrupt network predictions among newly generated perturbations. Hence, we do not consider all of the perturbations as treatments but selectively define them as worst perturbations $t$ breaking network predictions, such that it satisfies $y \neq f(x+t)$, where we call $x_\text{adv} = x + t$ as worst examples. This is because our major goal is to catch actual adversarial vulnerability of DNNs, so that we do not tackle the perturbations incapable of harming network predictions.

To access such worst perturbation, we choose perturbation generator $g$ as an adversarial attack of PGD according to standard adversarial training~\cite{madry2018towards}. In addition, we pick the worst perturbations $t$ damaging network predictions among adversarial perturbations from the generator $g$. In this way, we perform adversarial data generating process.

\subsection{Adversarial Problem Setup}

In the nature of adversarial training, the worst perturbations $t$ are explicitly injected to clean images $x$ such that $x_\text{adv}=x+t$, and these combined images are propagated into DNNs $f$. Here, through this formulation as: $f(x+t)=f(x,t)$, we connect DNNs for adversarial training and a nuisance parameter $f$ for non-parametric DML in Eq.~(\ref{eqn:np_dml}). Fortunately, once we use Taylor expansion (with scalar-valued function for better understanding) and decompose $f$ by its input component as: $f(x+t)=f(x)+\sum_{i=1}^{\infty} t^if^{(i)}(x)/i!$, where $f^{(i)}$ indicates $i$-th order derivative function, we can also express partially linear settings described in Eq.~(\ref{eqn:partially}). That is, since adversarial examples start from the concept of ``additive noise'', both settings can exist at the same time in the scheme of adversarial training. From this reason, we build \textit{Adversarial Double Machine Learning (ADML)}:
\begin{equation}
\label{eqn:adml_setup}
\begin{split}
       y &= f(x+t)=f(x) + \theta \bar{t}+ u, \quad (\E[u\mid x, t]=0)\\
       t &= g(x) + v, \quad\quad\quad\quad\quad\quad\quad\quad (\E[v \mid x]=0)
\end{split}
\end{equation}
where $\overline{t}$ indicates Taylor-order matrix: {\small $[t,t^2, \cdots]^T$} and $\theta$ represents Taylor-coefficient matrix {\small $[\frac{f^{(1)}(x)}{1!}, \frac{f^{(2)}(x)}{2!}, \cdots]$} (see strict mathematical verification in Appendix B).

Here, we explain what the conditional expected error of $u$ and $v$ in Eq.~(\ref{eqn:adml_setup}) means in adversarial training. The former $\E[u|x,t]=0$ implies the nature of adversarial training, which can be viewed as an ultimate augmentation robustifying DNNs, when infinite data population of $x$ and $t$ is given. Thus, it means network predictions become invariant in the end, despite the given worst perturbations. To implement it practically, we replace it with a mild assumption as $\E[u| x, g(x)]=0$ (see Appendix C) that signifies PGD-based perturbations are used to perform adversarial training. This is because we cannot always acquire worst perturbations $t$ using only PGD. For the latter $\E[v| x]=0$, it represents PGD has capability of producing worst perturbations deviating network predictions during the training.

\subsection{Estimating Adversarially Causal Parameter}

Aligned with Eq.~(\ref{eqn:partially}), an explicit term of $\theta$ is regarded as the causal parameter in our problem setup of ADML. We can now interpret that $\theta$  is a causal factor to spur adversarial vulnerability, since its magnitude easily catalyzes the deviations from network predictions of clean images. Here, if it is possible to directly compute $\theta$ over all data samples, we can finally handle adversarial vulnerability.

Favorably, ADML follows both partially linear and non-parametric settings due to the concept of additive noise, thus we can employ the way of estimating causal parameter as in Eq.~(\ref{eqn:nl_theta}). The following formulation represents the estimated causal parameter  $\hat{\theta}$ in ADML (see Appendix D). Note that, as we emphasized, \textit{sample-splitting} plus \textit{cross-fitting} must be applied to estimate unbiased causal parameter.
\begin{equation}
\label{eqn:adml_theta}
       \hat{\theta} =\E_{\mathcal{D}_t}\left[-\left(\frac{1}{p(\mathcal{T}{=}t \mid x)}-1\right)\frac{\partial}{\partial t}f(x+t)\right],
\end{equation}
where $\frac{\partial}{\partial t}f(x+t)$ indicates an input gradient for network predictions with respect to $t$, and $p(\mathcal{T}{=}t|x)$ represents a distribution of worst perturbation given clean images $x$. Here, we cannot directly handle this distribution due to the presence of multiple unknown parameters required to define it. For that reason, we instead approximate it with the sharpening technique by incorporating the information on attacked confidence such that $p(\mathcal{T}{=}t|x)\approx \E_{t'|x}[p(y_{a}|x, t')]$ (see Appendix E), where $y_{a}$ denotes attacked classes for the given worst perturbations $t$. It implicitly means that the higher the attacked confidence, the higher the probability of finding worst perturbations.

Aligned with the previous analysis~\cite{simon2019first} that show increasing magnitude of input gradient increases adversarial vulnerability, the magnitude of our causal parameter $|\hat{\theta}|$ also becomes huge due to $|\hat{\theta}| \propto |\frac{\partial}{\partial t}f(x+t)|$. In parallel, Qin \etal~\cite{qin2021improving} show the more ambiguous confident, the lower robustness (high vulnerability), and interestingly, the magnitude of our causal parameter also $|\hat{\theta}|$ becomes large due to $|\hat{\theta}
| \propto |1/\E_{t'|x}[p(y_{a}|x, t')]|$. 



Bringing such factors at once, $\hat{\theta}$ represents a weighted measurement of attacked confidence and their input gradients. Comprehensively, we can revisit that the network predictions of worst examples are easily flipped due to the following adversarial vulnerability: (a) ambiguous confidence around classification boundaries, or (b) high gradient magnitude amplifying the leverage of the perturbations. To improve the adversarial robustness of DNNs, it is essential to minimize the negative effects of causal parameters, which are combinatorial outcomes of the gradient and confidence.

\setlength{\textfloatsep}{5pt} 
\begin{algorithm}[t!]
\caption{ADML}
\label{alg:1}
\begin{algorithmic}[1]
\Require Data Samples $\mathcal{D}$, Network $f$
\For {$(x, y)\sim\mathcal{D}$} \Comment{Cross-Fitting}
\State $t' \gets g(x)$ \Comment{PGD Attack}
\State $(x_1, y_1, t_1'), (x_2, y_2, t_2') \sim \text{Split}(x, y, t')$
\State $\mathcal{L}_{a}\leftarrow\mathcal{L}_\text{Defense}(x_1, y_1, t_1';f)$ \Comment{Mild Assumption}
\State $(x_{t_{2}}, y_{t_{2}}, t_{2})\leftarrow \text{Select}(y_2 \neq f(x_2+t_2'
))$ \Comment{Worst}
\State $j^*\leftarrow\arg\max_j f_j(x_{t_{2}}+t_{2})$ \Comment{$j:$ Class Index}
\State $\tau \leftarrow \frac{1}{f_{j^*}(x_{t_{2}}+t_{2})}-1$ \Comment{Balancing Ratio}
\State $\mathcal{L}_{b} \leftarrow \tau\mathcal{L}_{\text{CE}}(f(x_{t_{2}}+t_{2}), y_{t_{2}})+\mathcal{L}_{\text{CE}}(f(x_{t_{2}}), y_{t_{2}})$
\State $\mathcal{L}_{\text{ADML}}\leftarrow\mathcal{L}_{a}+\mathcal{L}_{b}$ \Comment{ADML Loss}
\State $w_f \leftarrow w_f-\alpha\frac{\partial}{\partial w_f}\mathcal{L}_{\text{ADML}}$ \Comment{Weight Update ($\alpha$: lr)}
\EndFor
\end{algorithmic}
\end{algorithm}

\begin{table*}[t!]
\vspace{-0.8cm}
\centering
\renewcommand{\tabcolsep}{2.0mm}
\resizebox{0.96\linewidth}{!}{
\begin{tabular}{clccccccccccccccccccccc}
\Xhline{3\arrayrulewidth}\rule{0pt}{9pt}
                           & \multirow{2}{*}{Method} & \multicolumn{7}{c}{CIFAR-10}                                                                                  & \multicolumn{7}{c}{CIFAR-100}                                                                                 & \multicolumn{7}{c}{Tiny-ImageNet}                                                                              \\
\cmidrule(lr){3-9}\cmidrule(lr){10-16}\cmidrule(lr){17-23}
                           &                         & Clean         & BIM           & PGD           & CW$_{\infty}$            & AP            & DLR           & AA            & Clean         & BIM           & PGD           & CW$_{\infty}$            & AP            & DLR           & AA            & Clean         & BIM           & PGD           & CW$_{\infty}$            & AP            & DLR           & AA             \\
\midrule
\multirow{8}{*}{\rotatebox[origin=c]{90}{VGG-16}}    & AT                     & 78.8          & 49.4          & 48.1          & 46.8          & 46.4          & 46.4          & 46.3          & \textbf{53.9} & 26.0          & 25.0          & 24.1          & 23.7          & 23.8          & 23.7          & \textbf{56.5} & 26.6         & 25.4          & 25.5          & 24.7          & 24.6          & 24.6           \\
                           & +ADML                    & \textbf{80.9} & \textbf{61.8} & \textbf{61.7} & \textbf{59.8} & \textbf{55.0} & \textbf{54.8} & \textbf{54.5} & 52.2          & \textbf{31.0} & \textbf{30.8} & \textbf{29.9} & \textbf{27.8} & \textbf{27.6} & \textbf{27.3} & 55.4          & \textbf{35.5} & \textbf{34.9} & \textbf{32.9} & \textbf{32.7} & \textbf{32.4} & \textbf{32.2}  \\
\cmidrule{2-23}
                           & TRADES                  & 79.5          & 48.6          & 47.6          & 45.7          & 46.4          & 46.3          & 46.3          & \textbf{53.3} & 25.5          & 24.8          & 23.5          & 23.6          & 23.7          & 23.2          & \textbf{56.1} & 28.3          & 27.3          & 26.2          & 25.8          & 25.8          & 25.7           \\
                           & +ADML                    & \textbf{81.0} & \textbf{62.6} & \textbf{62.4} & \textbf{59.0} & \textbf{55.3} & \textbf{55.1} & \textbf{54.9} & 52.2          & \textbf{31.2} & \textbf{31.1} & \textbf{29.8} & \textbf{27.6} & \textbf{27.3} & \textbf{27.3} & 55.4          & \textbf{36.6} & \textbf{36.0} & \textbf{34.0} & \textbf{33.6} & \textbf{33.4} & \textbf{33.3}  \\
\cmidrule{2-23}
                           & MART                    & 78.3          & 51.9          & 50.6          & 48.8          & 48.9          & 48.8          & 48.7          & \textbf{52.6} & 26.6          & 26.0          & 24.4          & 24.3          & 24.3          & 24.2          & \textbf{55.7} & 27.8          & 26.6          & 26.0          & 25.7          & 25.7          & 25.6           \\
                           & +ADML                    & \textbf{80.4} & \textbf{62.4} & \textbf{62.2} & \textbf{60.3} & \textbf{55.7} & \textbf{55.3} & \textbf{55.2} & 51.6          & \textbf{31.6} & \textbf{31.0} & \textbf{29.6} & \textbf{27.8} & \textbf{27.5} & \textbf{27.3} & 55.1          & \textbf{36.2} & \textbf{35.8} & \textbf{34.9} & \textbf{34.3} & \textbf{33.9} & \textbf{33.7}  \\
\cmidrule{2-23}
                           & AWP                     & 77.2          & 53.9          & 52.6          & 50.1          & 51.4          & 51.1          & 51.0          & 52.1          & 30.2          & 29.3          & 27.5          & 28.7          & 28.5          & 28.3          & \textbf{56.9} & 31.6          & 31.0          & 29.4          & 29.9          & 29.8          & 29.7           \\
                           & +ADML                    & \textbf{80.2} & \textbf{64.7} & \textbf{64.6} & \textbf{61.8} & \textbf{58.0} & \textbf{57.7} & \textbf{57.5} & \textbf{52.3} & \textbf{33.8} & \textbf{33.5} & \textbf{31.3} & \textbf{29.7} & \textbf{29.5} & \textbf{29.0} & 54.5          & \textbf{36.6} & \textbf{36.1} & \textbf{34.6} & \textbf{34.5} & \textbf{33.8} & \textbf{33.7}  \\
\midrule
\multirow{8}{*}{\rotatebox[origin=c]{90}{ResNet-18}} & AT                     & 83.1          & 53.3          & 51.9          & 50.8          & 49.9          & 49.7          & 49.5          & \textbf{59.1} & 27.1          & 26.3          & 25.4          & 25.3          & 25.3          & 25.1          & \textbf{61.5} & 31.0          & 30.1          & 29.5          & 28.8          & 28.9          & 28.8           \\
                           & +ADML                    & \textbf{84.5} & \textbf{61.1} & \textbf{60.8} & \textbf{58.5} & \textbf{56.7} & \textbf{56.2} & \textbf{55.6} & 56.4          & \textbf{31.6} & \textbf{30.9} & \textbf{29.5} & \textbf{28.4} & \textbf{28.2} & \textbf{27.8} & 57.3          & \textbf{35.9} & \textbf{35.5} & \textbf{33.5} & \textbf{34.7} & \textbf{34.7} & \textbf{34.6}  \\
\cmidrule{2-23}
                           & TRADES                  & 83.3          & 53.0          & 52.0          & 50.9          & 50.9          & 50.8          & 50.7          & \textbf{58.5} & 27.6          & 26.8          & 26.2          & 25.9          & 25.9          & 25.8          & \textbf{60.4} & 32.0          & 31.0          & 30.2          & 29.7          & 29.6          & 29.5           \\
                           & +ADML                    & \textbf{84.0} & \textbf{62.3} & \textbf{61.9} & \textbf{59.5} & \textbf{56.7} & \textbf{56.6} & \textbf{55.9} & 57.3          & \textbf{31.7} & \textbf{31.6} & \textbf{30.0} & \textbf{28.8} & \textbf{28.5} & \textbf{28.0} & 58.4          & \textbf{37.0} & \textbf{36.0} & \textbf{33.3} & \textbf{35.0} & \textbf{34.1} & \textbf{34.0}  \\
\cmidrule{2-23}
                           & MART                    & 82.5          & 54.1          & 52.8          & 51.3          & 51.5          & 50.8          & 50.8          & \textbf{58.1} & 27.7          & 26.7          & 25.5          & 25.4          & 25.1          & 25.0          & \textbf{60.6} & 31.0          & 30.2          & 29.8          & 29.0          & 29.0          & 29.0           \\
                           & +ADML                    & \textbf{84.1} & \textbf{63.3} & \textbf{62.9} & \textbf{58.7} & \textbf{56.8} & \textbf{56.4} & \textbf{56.2} & 57.4          & \textbf{32.1} & \textbf{31.7} & \textbf{30.2} & \textbf{28.9} & \textbf{28.8} & \textbf{28.5} & 57.1          & \textbf{36.6} & \textbf{36.0} & \textbf{33.6} & \textbf{35.1} & \textbf{33.9} & \textbf{33.8}  \\
\cmidrule{2-23}
                           & AWP                     & 81.3          & 56.3          & 55.5          & 53.6          & 54.2          & 54.0          & 54.0          & \textbf{57.9} & 31.4          & 30.7          & 29.0          & 30.0          & 30.0          & 29.8          & \textbf{61.4} & 34.7          & 34.1          & 32.4          & 33.3          & 33.2          & 33.1           \\
                           & +ADML                    & \textbf{84.0} & \textbf{64.6} & \textbf{64.5} & \textbf{61.4} & \textbf{60.5} & \textbf{59.9} & \textbf{59.7} & 56.2          & \textbf{33.9} & \textbf{32.9} & \textbf{30.7} & \textbf{31.1} & \textbf{30.5} & \textbf{30.3} & 59.8          & \textbf{38.6} & \textbf{38.1} & \textbf{35.9} & \textbf{37.7} & \textbf{36.7} & \textbf{36.6}  \\
\midrule
\multirow{8}{*}{\rotatebox[origin=c]{90}{WideResNet-28-10}} & AT                     & 86.7          & 55.4          & 53.4          & 53.4          & 51.3          & 51.3          & 51.2          & \textbf{61.9} & 28.8          & 27.4          & 27.1          & 26.0          & 26.0          & 25.9          & \textbf{64.8} & 32.7          & 31.2          & 31.1          & 30.1          & 30.1          & 30.0          \\
                           & +ADML                    & \textbf{87.5} & \textbf{61.7} & \textbf{60.7} & \textbf{58.8} & \textbf{56.4} & \textbf{56.3} & \textbf{55.8} & 58.9          & \textbf{32.9} & \textbf{32.6} & \textbf{31.3} & \textbf{29.6} & \textbf{29.2} & \textbf{29.1} & 62.1          & \textbf{43.5} & \textbf{43.1} & \textbf{41.1} & \textbf{41.9} & \textbf{40.4} & \textbf{40.2} \\
\cmidrule{2-23}
                           & TRADES                  & 86.0          & 55.3          & 53.7          & 53.6          & 51.6          & 51.6          & 51.4          & \textbf{61.9} & 29.1          & 28.5          & 27.8          & 26.7          & 26.8          & 26.7          & \textbf{64.2} & 32.5          & 31.6          & 31.4          & 30.0          & 29.9          & 29.8          \\
                           & +ADML                    & \textbf{88.5} & \textbf{62.9} & \textbf{61.9} & \textbf{59.6} & \textbf{57.6} & \textbf{57.6} & \textbf{56.6} & 61.6          & \textbf{33.3} & \textbf{33.0} & \textbf{31.5} & \textbf{30.0} & \textbf{30.1} & \textbf{29.6} & 63.1          & \textbf{43.5} & \textbf{42.9} & \textbf{41.0} & \textbf{41.4} & \textbf{40.3} & \textbf{40.2} \\
\cmidrule{2-23}
                           & MART                    & 86.4          & 56.0          & 54.3          & 53.4          & 51.6          & 51.6          & 51.5          & \textbf{61.6} & 28.3          & 26.7          & 26.4          & 25.3          & 25.4          & 25.3          & \textbf{64.2} & 32.9          & 31.8          & 31.8          & 30.7          & 30.5          & 30.5          \\
                           & +ADML                    & \textbf{88.3} & \textbf{62.1} & \textbf{60.9} & \textbf{59.6} & \textbf{56.7} & \textbf{56.6} & \textbf{56.2} & 59.6          & \textbf{33.0} & \textbf{32.9} & \textbf{31.6} & \textbf{30.3} & \textbf{29.9} & \textbf{29.6} & 61.8          & \textbf{42.8} & \textbf{42.6} & \textbf{40.3} & \textbf{40.7} & \textbf{39.0} & \textbf{38.9} \\
\cmidrule{2-23}
                           & AWP                     & 85.9          & 60.2          & 58.9          & 57.2          & 56.9          & 56.9          & 56.8          & 62.4          & 33.0          & 32.2          & 31.1          & 30.9          & 30.9          & 30.8          & \textbf{65.4} & 36.9          & 36.0          & 35.1          & 34.8          & 34.8          & 34.7          \\
                           & +ADML                    & \textbf{88.2} & \textbf{67.5} & \textbf{67.4} & \textbf{64.2} & \textbf{63.5} & \textbf{63.2} & \textbf{63.1} & \textbf{62.5} & \textbf{39.7} & \textbf{39.3} & \textbf{36.9} & \textbf{37.6} & \textbf{37.1} & \textbf{36.8} & 64.9          & \textbf{44.6} & \textbf{44.2} & \textbf{41.9} & \textbf{43.3} & \textbf{42.1} & \textbf{42.0} \\
\midrule
\multirow{8}{*}{\rotatebox[origin=c]{90}{WideResNet-70-10}} & AT                     & 88.1          & 56.6          & 54.8          & 55.0          & 52.8          & 53.0          & 52.8          & \textbf{64.1} & 28.4          & 27.3          & 27.4          & 26.0          & 26.4          & 25.6          & \textbf{65.3} & 34.9          & 33.4          & 33.9          & 32.2          & 32.2          & 32.1           \\
                           & +ADML                    & \textbf{88.9} & \textbf{61.5} & \textbf{61.4} & \textbf{61.0} & \textbf{56.5} & \textbf{56.3} & \textbf{56.0} & 63.3          & \textbf{30.0} & \textbf{29.2} & \textbf{28.8} & \textbf{26.9} & \textbf{26.7} & \textbf{26.4} & 61.0          & \textbf{37.7} & \textbf{37.4} & \textbf{36.9} & \textbf{33.8} & \textbf{33.0} & \textbf{32.9}  \\
\cmidrule{2-23}
                           & TRADES                  & 87.7          & 56.3          & 54.7          & 55.0          & 53.4          & 53.3          & 53.3          & 63.3          & 28.7          & 27.8          & 27.9          & 26.6          & 26.2          & 26.0          & \textbf{65.7} & 34.4          & 32.6          & 33.0          & 31.5          & 31.5          & 31.4           \\
                           & +ADML                    & \textbf{89.1} & \textbf{63.9} & \textbf{63.3} & \textbf{62.7} & \textbf{59.0} & \textbf{59.6} & \textbf{59.0} & \textbf{63.4} & \textbf{31.2} & \textbf{30.8} & \textbf{30.3} & \textbf{27.5} & \textbf{27.1} & \textbf{27.0} & 61.8          & \textbf{40.2} & \textbf{39.5} & \textbf{38.8} & \textbf{36.1} & \textbf{35.5} & \textbf{35.4}  \\
\cmidrule{2-23}
                           & MART                    & 88.0          & 57.4          & 55.5          & 55.4          & 52.8          & 52.8          & 52.6          & \textbf{63.2} & 28.7          & 27.5          & 27.5          & 25.8          & 26.3          & 25.6          & \textbf{65.4} & 33.8          & 32.5          & 32.4          & 31.3          & 31.3          & 31.2           \\
                           & +ADML                    & \textbf{88.5} & \textbf{61.7} & \textbf{61.3} & \textbf{60.8} & \textbf{56.7} & \textbf{56.8} & \textbf{56.6} & 62.3          & \textbf{30.1} & \textbf{30.0} & \textbf{29.4} & \textbf{29.3} & \textbf{29.1} & \textbf{28.7} & 63.2          & \textbf{41.8} & \textbf{41.0} & \textbf{40.2} & \textbf{37.7} & \textbf{36.6} & \textbf{36.5}  \\
\cmidrule{2-23}
                           & AWP                     & 86.6          & 61.8          & 60.6          & 59.9          & 59.1          & 59.4          & 59.2          & 65.2          & 33.3          & 33.3          & 31.7          & 31.5          & 30.3          & 30.0          & \textbf{66.7} & 40.7          & 40.0          & 40.0          & 39.1          & 39.0          & 38.9           \\
                           & +ADML                    & \textbf{89.4} & \textbf{67.0} & \textbf{66.9} & \textbf{66.1} & \textbf{63.4} & \textbf{63.6} & \textbf{63.1} & \textbf{65.3} & \textbf{41.9} & \textbf{41.8} & \textbf{40.9} & \textbf{38.9} & \textbf{38.0} & \textbf{37.6} & 65.8          & \textbf{45.0} & \textbf{44.5} & \textbf{43.3} & \textbf{43.5} & \textbf{43.1} & \textbf{42.8} \\
                           \Xhline{3\arrayrulewidth}\rule{0pt}{9pt}
\end{tabular}
}
\vspace{-0.2cm}
\caption{Comparing adversarial robustness of various defence methods whether to the inclusion of ADML for CIFAR-10~\cite{krizhevsky2009learning}, CIFAR-100~\cite{krizhevsky2009learning}, Tiny-ImageNet~\cite{le2015tiny} trained with VGG-16~\cite{vgg}, ResNet-18~\cite{resnet}, WideResNet-28-10~\cite{wideresent}, and WideResNet-70-10~\cite{wideresent}.}
\vspace{-0.3cm}
\label{table:cnn}
\end{table*}

\subsection{Mitigating Adversarial Vulnerability}

By deploying ADML, we propose a way of estimating causal parameter representing the degree of adversarial vulnerability that disturbs to predict the target classes. Then, our final goal is essentially to lessen its direct causal effect from adversarial perturbations in order to achieve robust networks. In detailed, alleviating their causal effect derived from $\hat{\theta}$ is the process of comprehensive reconstruction to focus more on vulnerable samples as we reflect their attacked confidence and gradients effects altogether.

Accordingly, the very first way is naively reducing the magnitude of $\hat{\theta}$ to suppress adversarial vulnerability damaging the robustness. However, calculating $\hat{\theta}$ and minimizing its magnitude at every iteration is computationally striking because input gradient has huge dimension of $\mathbb{R}^{dhwc}$ and getting its gradient inevitably needs to compute second-order gradient with its tremendous dimension. We instead approximate the partial derivative $\frac{\partial}{\partial t}\E[y|\text{do}(\mathcal{T}{=}t)]$ and minimize its magnitude, which can be written as:
\begin{equation}
\label{eqn:approx_theta}
    \min_f|\hat{\theta}|\approx\left|\frac{\E[y\mid\text{do}(\mathcal{T}{=}t)]-\E[y\mid\text{do}(\mathcal{T}{=}0)]}{t-0}\right|,
\end{equation}
where network parameters of DNNs $f$ are only dependent on the numerator, thus we engross the numerator only. Lastly, we redesign $\E[y|\text{do}(\mathcal{T}{=}t)]$ into the form of loss function used in deep learning and finally construct the objective function for ADML, of which formulation can be written as follows (see details in Appendix F):
\begin{equation}
\label{eqn:final_adml_objective}
    \min_{f} \E_{\mathcal{D}_t}[ \tau\mathcal{L}_{\text{CE}}(f(x+t), y)]+ \E_{\mathcal{D}_0}[\mathcal{L}_{\text{CE}}(f(x), y)],
\end{equation}
where we denote {$\tau = \frac{1}{p(\mathcal{T}{=}t\mid x)}-1$} as balancing ratio. The current AT-based defenses use an equal weight ``$1/n$'' to loss for all data samples: {$\frac{1}{n}\sum_{i{=}1}^{n}\mathcal{L}_{\text{Defense}}(x_i,y_i,t_i;f)$} because they presume all of perturbations have equal causal effect (successful attack) to change targets without realizing vulnerable samples. Whereas, ADML uses the balancing ratio $\tau$ to adaptively focus on vulnerable samples by reweighting the loss. To realize ADML, we describe Algorithm~\ref{alg:1} to explain further details, where $\mathcal{L}_{\text{Defense}}(x,y,t;f)$ indicates a main body of the loss function for AT-based defenses.


\begin{table*}[t!]
\vspace{-0.8cm}
\centering
\renewcommand{\tabcolsep}{2.0mm}
\resizebox{0.96\linewidth}{!}{
\begin{tabular}{clccccccccccccccccccccc}
\Xhline{3\arrayrulewidth}\rule{0pt}{9pt}
                           & \multirow{2}{*}{Method} & \multicolumn{7}{c}{CIFAR-10}                                                                                  & \multicolumn{7}{c}{CIFAR-100}                                                                                 & \multicolumn{7}{c}{Tiny-ImageNet}                                                                              \\
\cmidrule(lr){3-9}\cmidrule(lr){10-16}\cmidrule(lr){17-23}
\multicolumn{1}{l}{}       & \multicolumn{1}{c}{}                                 & Clean & BIM  & PGD  & CW   & AP   & DLR  & AA   & Clean & BIM  & PGD  & CW$_{\infty}$   & AP   & DLR  & AA   & Clean & BIM  & PGD  & CW$_{\infty}$   & AP   & DLR  & AA   \\
\midrule
\multirow{8}{*}{\rotatebox[origin=c]{90}{ViT-S/16}}  & AT                                                  & 83.5           & 49.9          & 47.3          & 46.6          & 44.9          & 44.8          & 44.7          & 59.9           & 26.6          & 25.8          & 25.1          & 24.6          & 24.5          & 24.5          & 73.8           & 35.8          & 34.1          & 33.5          & 31.9          & 31.9          & 31.8          \\
                           & +ADML                                                 & \textbf{88.1}  & \textbf{56.8} & \textbf{55.1} & \textbf{53.8} & \textbf{51.3} & \textbf{50.7} & \textbf{50.7} & \textbf{62.7}  & \textbf{32.4} & \textbf{30.7} & \textbf{29.4} & \textbf{28.4} & \textbf{27.9} & \textbf{27.7} & \textbf{75.6}  & \textbf{47.7} & \textbf{46.6} & \textbf{45.6} & \textbf{45.3} & \textbf{42.7} & \textbf{42.6} \\
\cmidrule{2-23}
                           & TRADES                                               & 85.0           & 51.0          & 49.4          & 48.6          & 48.1          & 48.0          & 47.8          & 59.5           & 27.3          & 26.6          & 26.3          & \textbf{25.8} & 25.9          & 25.7          & 72.9           & 38.8          & 37.8          & 37.4          & 36.1          & 36.1          & 36.0          \\
                           & +ADML                                                 & \textbf{87.9}  & \textbf{57.6} & \textbf{56.2} & \textbf{55.1} & \textbf{52.7} & \textbf{52.0} & \textbf{51.9} & \textbf{63.2}  & \textbf{35.0} & \textbf{34.7} & \textbf{33.8} & \textbf{31.6} & \textbf{31.3} & \textbf{31.2} & \textbf{75.3}  & \textbf{49.1} & \textbf{48.0} & \textbf{47.1} & \textbf{45.6} & \textbf{43.0} & \textbf{43.0} \\
\cmidrule{2-23}
                           & MART                                                 & 85.7           & 52.4          & 49.7          & 48.9          & 46.7          & 46.7          & 46.6          & 60.9           & 28.6          & 27.9          & 27.5          & 26.5          & 26.5          & 26.4          & \textbf{77.6}  & 38.6          & 37.2          & 36.8          & 35.2          & 35.2          & 35.2          \\
                           & +ADML                                                 & \textbf{88.0}  & \textbf{57.5} & \textbf{56.1} & \textbf{54.7} & \textbf{51.9} & \textbf{51.4} & \textbf{51.4} & \textbf{62.7}  & \textbf{34.4} & \textbf{32.7} & \textbf{31.7} & \textbf{30.0} & \textbf{29.3} & \textbf{29.2} & 76.5  & \textbf{48.9} & \textbf{47.7} & \textbf{46.3} & \textbf{46.0} & \textbf{42.9} & \textbf{42.9} \\
\cmidrule{2-23}
                           & AWP                                                  & 84.9           & 54.2          & 52.3          & 51.6          & \textbf{50.1} & \textbf{49.9} & \textbf{49.9} & 61.1           & 29.6          & 29.0          & 28.1          & 27.7          & 27.7          & 27.6          & \textbf{78.1}  & 41.0          & 39.3          & 38.7          & 37.3          & 37.3          & 37.2          \\
                           & +ADML                                                 & \textbf{88.0}  & \textbf{57.0} & \textbf{54.3} & \textbf{53.3} & \textbf{50.1} & 49.6          & 49.5          & \textbf{64.4}  & \textbf{34.8} & \textbf{34.1} & \textbf{33.0} & \textbf{31.3} & \textbf{30.6} & \textbf{30.5} & 76.6  & \textbf{47.5} & \textbf{46.4} & \textbf{44.9} & \textbf{44.7} & \textbf{42.5} & \textbf{42.4} \\
\midrule
\multirow{8}{*}{\rotatebox[origin=c]{90}{ViT-B/16}}  & AT                                                  & 87.0           & 52.8          & 50.8          & 50.4          & 47.8          & 47.7          & 47.7          & 63.3           & 30.4          & 29.6          & 29.2          & 28.6          & 28.3          & 28.3          & 72.4           & 40.1          & 37.7          & 37.8          & 34.4          & 34.3          & 34.5          \\
                           & +ADML                                                 & \textbf{89.9}  & \textbf{56.1} & \textbf{54.9} & \textbf{54.1} & \textbf{51.6} & \textbf{51.4} & \textbf{51.2} & \textbf{67.1}  & \textbf{38.1} & \textbf{36.1} & \textbf{35.4} & \textbf{34.3} & \textbf{33.4} & \textbf{33.1} & \textbf{79.0}  & \textbf{50.2} & \textbf{49.7} & \textbf{48.4} & \textbf{48.5} & \textbf{46.9} & \textbf{46.8} \\
\cmidrule{2-23}
                           & TRADES                                               & 85.3           & 53.8          & 52.4          & 51.6          & 50.9          & 50.8          & 50.8          & 65.7           & 32.6          & 31.5          & 31.0          & 29.9          & 30.0          & 30.0          & 73.2           & 43.3          & 41.2          & 41.8          & 39.1          & 39.0          & 39.4          \\
                           & +ADML                                                 & \textbf{88.9}  & \textbf{58.6} & \textbf{57.3} & \textbf{56.1} & \textbf{54.7} & \textbf{54.4} & \textbf{54.3} & \textbf{69.4}  & \textbf{38.9} & \textbf{37.9} & \textbf{37.0} & \textbf{34.8} & \textbf{34.6} & \textbf{34.7} & \textbf{79.4}  & \textbf{54.0} & \textbf{52.2} & \textbf{51.6} & \textbf{48.2} & \textbf{47.3} & \textbf{47.2} \\
\cmidrule{2-23}
                           & MART                                                 & 87.4           & 53.3          & 50.6          & 50.5          & 48.3          & 48.4          & 48.2          & 65.7           & 31.9          & 30.8          & 30.2          & 29.2          & 29.2          & 29.1          & 79.3           & 41.7          & 40.0          & 39.6          & 36.8          & 36.8          & 37.1          \\
                           & +ADML                                                 & \textbf{89.6}  & \textbf{57.0} & \textbf{55.6} & \textbf{54.3} & \textbf{51.8} & \textbf{51.5} & \textbf{51.3} & \textbf{68.9}  & \textbf{35.4} & \textbf{33.5} & \textbf{32.9} & \textbf{30.5} & \textbf{30.1} & \textbf{30.2} & \textbf{80.1}  & \textbf{50.7} & \textbf{50.1} & \textbf{49.0} & \textbf{48.9} & \textbf{47.4} & \textbf{47.0} \\
\cmidrule{2-23}
                           & AWP                                                  & 87.4           & 54.9          & 52.9          & 51.9          & 49.8          & 49.8          & 49.7          & 66.8           & 33.4          & 31.7          & 31.3          & 30.3          & 30.2          & 30.2          & 78.7           & 45.0          & 42.2          & 42.2          & 39.5          & 39.4          & 39.7          \\
                           & +ADML                                                 & \textbf{90.4}  & \textbf{59.1} & \textbf{56.7} & \textbf{56.6} & \textbf{53.8} & \textbf{53.6} & \textbf{52.8} & \textbf{70.0}  & \textbf{37.7} & \textbf{36.1} & \textbf{35.6} & 33.3          & \textbf{32.8} & \textbf{32.9} & \textbf{79.3}  & \textbf{50.8} & \textbf{49.7} & \textbf{48.0} & \textbf{48.0} & \textbf{46.0} & \textbf{45.8} \\
\midrule
\multirow{8}{*}{\rotatebox[origin=c]{90}{DeiT-S/16}} & AT                                                  & 83.5           & 49.3          & 47.8          & 46.7          & 45.3          & 45.3          & 45.2          & 59.5           & 29.2          & 28.5          & 27.6          & 27.5          & 27.5          & 27.4          & \textbf{75.7}  & 37.4          & 35.6          & 34.7          & 33.1          & 33.0          & 33.0          \\
                           & +ADML                                                 & \textbf{87.7}  & \textbf{56.8} & \textbf{56.0} & \textbf{54.9} & \textbf{52.3} & \textbf{51.9} & \textbf{51.9} & \textbf{63.7}  & \textbf{34.4} & \textbf{32.9} & \textbf{31.7} & \textbf{31.4} & \textbf{30.5} & \textbf{30.4} & 74.7           & \textbf{44.9} & \textbf{43.6} & \textbf{42.2} & \textbf{40.8} & \textbf{39.5} & \textbf{39.4} \\
\cmidrule{2-23}
                           & TRADES                                               & 84.1           & 50.6          & 49.3          & 48.8          & 48.0          & 48.0          & 48.0          & 61.8           & 29.4          & 28.8          & 27.8          & 28.1          & 28.0          & 28.0          & 74.8  & 39.0          & 38.0          & 37.4          & 36.4          & 36.4          & 36.3          \\
                           & +ADML                                                 & \textbf{87.9}  & \textbf{57.8} & \textbf{56.5} & \textbf{55.3} & \textbf{53.7} & \textbf{53.0} & \textbf{53.2} & \textbf{66.2}  & \textbf{37.2} & \textbf{36.4} & \textbf{35.5} & \textbf{33.2} & \textbf{32.3} & \textbf{32.3} & \textbf{76.3}  & \textbf{45.2} & \textbf{44.9} & \textbf{43.8} & \textbf{39.5} & \textbf{38.7} & \textbf{38.4} \\
\cmidrule{2-23}
                           & MART                                                 & 84.2           & 52.3          & 50.0          & 49.1          & 47.8          & 47.6          & 47.5          & 59.8           & 31.0          & 30.6          & 29.3          & 29.7          & 29.6          & 29.6          & 74.6  & 40.1          & 39.1          & 38.4          & 37.7          & 37.6          & 37.6          \\
                           & +ADML                                                 & \textbf{87.5}  & \textbf{57.5} & \textbf{55.6} & \textbf{55.0} & \textbf{52.6} & \textbf{52.3} & \textbf{52.2} & \textbf{65.3}  & \textbf{37.0} & \textbf{35.6} & \textbf{34.7} & \textbf{32.4} & \textbf{30.7} & \textbf{30.7} & \textbf{75.2}  & \textbf{45.3} & \textbf{44.2} & \textbf{43.3} & \textbf{42.8} & \textbf{38.4} & \textbf{38.4} \\
\cmidrule{2-23}
                           & AWP                                                  & 82.3           & 53.5          & 52.3          & 51.5          & 50.5          & 50.4          & 50.4          & 60.7           & 31.8          & 31.4          & 30.2          & 31.0          & 30.0          & 30.0          & \textbf{75.4}  & 41.7          & 40.9          & 39.8          & 39.0          & 39.1          & 39.0          \\
                           & +ADML                                                 & \textbf{86.7}  & \textbf{55.9} & \textbf{53.2} & \textbf{52.6} & \textbf{50.6} & \textbf{50.5} & \textbf{50.5} & \textbf{64.7}  & \textbf{39.4} & \textbf{38.1} & \textbf{36.8} & \textbf{35.7} & \textbf{34.5} & \textbf{34.5} & \textbf{75.4}  & \textbf{49.4} & \textbf{47.6} & \textbf{46.5} & \textbf{45.7} & \textbf{42.8} & \textbf{42.8} \\
\midrule
\multirow{8}{*}{\rotatebox[origin=c]{90}{DeiT-B/16}} & AT                                                  & 84.6           & 51.5          & 49.5          & 48.4          & 47.2          & 47.1          & 47.0          & 64.9  & 30.3          & 29.1          & 28.4          & 27.5          & 27.4          & 27.4          & \textbf{79.1}  & 38.6          & 36.3          & 36.1          & 34.4          & 34.3          & 34.0          \\
                           & +ADML                                                 & \textbf{89.7}  & \textbf{55.7} & \textbf{54.8} & \textbf{53.8} & \textbf{50.0} & \textbf{49.7} & 49.7          & \textbf{65.7}  & \textbf{35.4} & \textbf{34.4} & \textbf{33.4} & \textbf{32.2} & \textbf{30.3} & \textbf{30.3} & 77.6           & \textbf{46.9} & \textbf{45.6} & \textbf{44.7} & \textbf{44.9} & \textbf{40.5} & \textbf{40.5} \\
                           \cmidrule{2-23}
                           & TRADES                                               & 85.4           & 52.8          & 51.7          & 50.6          & 50.2          & 50.2          & 50.2          & 64.8           & 30.0          & 29.3          & 28.6          & 28.4          & 28.3          & 28.3          & 78.3           & 43.1          & 41.6          & 40.5          & 40.5          & 40.5          & 40.4          \\
                           & +ADML                                                 & \textbf{90.2}  & \textbf{61.4} & \textbf{60.4} & \textbf{59.4} & \textbf{58.3} & \textbf{57.6} & \textbf{57.6} & \textbf{68.6}  & \textbf{40.5} & \textbf{39.9} & \textbf{38.4} & \textbf{37.4} & \textbf{36.8} & \textbf{36.7} & \textbf{80.8}  & \textbf{45.4} & \textbf{43.5} & \textbf{42.7} & \textbf{43.2} & \textbf{42.9} & \textbf{42.7} \\
                            \cmidrule{2-23}
                           & MART                                                 & 83.9           & 54.7          & 53.2          & 52.0          & 51.0          & 50.9          & 50.7          & 64.5           & 31.9          & 31.1          & 30.5          & 30.2          & 30.1          & 30.0          & 75.8  & 44.6          & 43.3          & 42.6          & 42.6          & 42.6          & 42.5          \\
                           & +ADML                                                 & \textbf{89.6}  & \textbf{60.3} & \textbf{60.2} & \textbf{58.9} & \textbf{55.1} & \textbf{55.0} & \textbf{55.0} & \textbf{65.3}  & \textbf{39.6} & \textbf{38.5} & \textbf{37.1} & \textbf{35.4} & \textbf{34.6} & \textbf{34.6} & \textbf{77.8}  & \textbf{47.7} & \textbf{46.1} & \textbf{44.6} & \textbf{45.5} & \textbf{44.8} & \textbf{44.7} \\
                            \cmidrule{2-23}
                           & AWP                                                  & 83.3           & 54.1          & 53.1          & 52.4          & 51.8          & 51.6          & 51.5          & 65.4           & 32.3          & 31.5          & 30.4          & 30.2          & 30.1          & 30.1          & 76.6           & 42.8          & 41.4          & 40.7          & 42.8          & 42.8          & 42.7          \\
                           & +ADML                                                 & \textbf{88.9}  & \textbf{59.0} & \textbf{57.3} & \textbf{56.4} & \textbf{54.0} & \textbf{53.8} & \textbf{53.7} & \textbf{69.7}  & \textbf{39.4} & \textbf{38.3} & \textbf{37.3} & \textbf{35.2} & \textbf{34.4} & \textbf{34.4} & \textbf{80.3}  & \textbf{51.2} & \textbf{50.2} & \textbf{48.9} & \textbf{49.4} & \textbf{48.1} & \textbf{48.0} \\
                            \Xhline{3\arrayrulewidth}\rule{0pt}{9pt}
\end{tabular}
}
\vspace{-0.2cm}
\caption{Comparing adversarial robustness of various defence methods whether to the inclusion of ADML for CIFAR-10~\cite{krizhevsky2009learning}, CIFAR-100~\cite{krizhevsky2009learning}, Tiny-ImageNet~\cite{le2015tiny} trained with ViT-S/16~\cite{dosovitskiy2020image}, ViT-B/16~\cite{dosovitskiy2020image}, DeiT-S/16~\cite{touvron2021training}, and DeiT-B/16~\cite{touvron2021training}.}
\vspace{-0.3cm}
\label{table:transformer}
\end{table*}

\section{Experiment}
\label{sec:experiment}

\subsection{Implementation Details}
\noindent\textbf{Datasets \& Networks.~} We conduct comprehensive experiments on various datasets and networks. For datasets, we use CIFAR-10~\cite{krizhevsky2009learning}, CIFAR-100~\cite{krizhevsky2009learning}, and two larger datasets: Tiny-ImageNet~\cite{le2015tiny} and ImageNet~\cite{deng2009imagenet}. For networks, four CNN architectures:~\cite{vgg,resnet, wideresent} and four Transformer architectures:~\cite{dosovitskiy2020image, touvron2021training} are used. 


\newcommand{\cmark}{\ding{51}}%
\newcommand{\xmark}{\ding{55}}%

\begin{table}[t!]
\centering
\renewcommand{\tabcolsep}{1.5mm}
\resizebox{0.99\linewidth}{!}{
\begin{tabular}{cccccccccccc}
\Xhline{3\arrayrulewidth}\rule{0pt}{9pt}
 & \multicolumn{1}{l}{} & \multicolumn{1}{l}{} & \multicolumn{1}{l}{} & \multicolumn{4}{c}{CIFAR-10}                                  & \multicolumn{4}{c}{Tiny-ImageNet}                             \\
\cmidrule(lr){5-8}\cmidrule(lr){9-12}
\multicolumn{1}{l}{}       & SS+CF                & Worst                & Non-Worst            & PGD           & CW$_{\infty}$            & DLR           & AA            & PGD           & CW$_{\infty}$            & DLR           & AA            \\
\midrule
\multirow{4}{*}{\rotatebox[origin=c]{90}{VGG-16}}    & \cmark    & \cmark    &   \xmark          & \textbf{61.7} & \textbf{59.8} & \textbf{54.8} & \textbf{54.4} & \textbf{34.9} & \textbf{32.9} & \textbf{32.4} & \textbf{32.2} \\
\cdashline{2-9}\noalign{\vskip 0.5ex}
                           & \cmark            & \cmark            & \cmark            & 52.3          & 49.4          & 48.9          & 48.8          & 25.4          & 24.9          & 24.2          & 24.2          \\
\cdashline{2-9}\noalign{\vskip 0.5ex}
                           & \cmark            &   \xmark         & \cmark            & 48.0          & 47.1          & 45.8          & 45.8          & 26.0          & 25.8          & 24.5          & 24.4          \\
\cdashline{2-9}\noalign{\vskip 0.5ex}
                           &    \xmark        & \cmark            &  \xmark           & 52.6          & 50.0          & 49.4          & 49.4          & 28.1          & 27.0          & 26.4          & 26.4          \\
\midrule
\multirow{4}{*}{\rotatebox[origin=c]{90}{ResNet-18}} & \cmark            & \cmark            &  \xmark        & \textbf{60.8} & \textbf{58.5} & \textbf{56.2} & \textbf{55.6} & \textbf{35.5} & \textbf{33.5} & \textbf{34.7} & \textbf{34.6} \\
\cdashline{2-9}\noalign{\vskip 0.5ex}
                           & \cmark            & \cmark            & \cmark            & 51.7          & 50.1          & 49.5          & 49.2          & 34.9          & 32.4          & 32.7          & 32.7          \\
\cdashline{2-9}\noalign{\vskip 0.5ex}
                           & \cmark            &     \xmark         & \cmark            & 50.8          & 50.5          & 49.4          & 49.1          & 30.0          & 29.5          & 29.0          & 28.9          \\
\cdashline{2-9}\noalign{\vskip 0.5ex}
                           &      \xmark        & \cmark            &   \xmark         & 53.6          & 51.7          & 51.4          & 51.0          & 31.7          & 29.7          & 30.2          & 30.1         \\
\Xhline{3\arrayrulewidth}\rule{0pt}{9pt}
\end{tabular}
}
\vspace{-0.2cm}
\caption{Ablation study for the effects of sample-splitting plus cross-fitting (SS+CF) and Worst/Non-Worst examples.}
\label{table:ablation}
\end{table}

\noindent\textbf{Adversarial Attacks.~} We adaptively set perturbation budget $\gamma$ of adversarial attacks depending on the classification difficulty of the four datasets: $8/255$ equally for CIFAR-10~\cite{krizhevsky2009learning} and CIFAR100~\cite{krizhevsky2009learning}, $4/255$ for Tiny-ImageNet~\cite{le2015tiny}, and $2/255$ for ImageNet~\cite{deng2009imagenet}. We prepare three standard attacks: BIM~\cite{45816}, PGD~\cite{madry2018towards}, CW$_{\infty}$~\cite{CW}, and three advanced attacks: AP (Auto-PGD: step size-free), DLR (Auto-DLR: shift and scaling invariant), AA (Auto-Attack: parameter-free), all of which are introduced by Francesco~\etal~\cite{croce2020reliable}. PGD, AP, DLR have $30$ steps with random starts, where PGD has step size $2.3\times\frac{\gamma}{30}$, and AP, DLR both have momentum coefficient $\rho=0.75$. CW$_{\infty}$ uses PGD-based gradient clamping for $l_{\infty}$ with CW objective~\cite{CW} on $\kappa=0$.

\noindent\textbf{Adversarial Defenses.~} We use four defense baselines with a standard baseline: AT~\cite{madry2018towards} and three advanced defense baselines: TRADES~\cite{pmlr-v97-zhang19p}, MART~\cite{Wang2020Improving}, AWP~\cite{wu2020adversarial}. To fairly validate experiments, a perturbation generator, PGD~\cite{madry2018towards} is equivalently used to generate adversarial examples for which we use the budget $8/255$ and set $10$ steps with $2.3\times\frac{\gamma}{10}$ step size in training. Especially, adversarially training Tiny-ImageNet~\cite{le2015tiny} and ImageNet~\cite{deng2009imagenet} is a computational burden, thus we employ fast adversarial training~\cite{Wong2020Fast} with FGSM~\cite{43405}. For training CNNs, we use SGD~\cite{ruder2016overview} with a learning rate of $0.5$ scheduled by Cyclic~\cite{smith2017cyclical} in $120$ epochs and use early stopping to 
prevent overfitting~\cite{pmlrv119rice20a}. For training Transformers, we use SGD~\cite{ruder2016overview} with a learning rate of $0.001$ on the equal experimental setup of CNNs, where $224\times224$ resolution is applied for all datasets and pretrained parameters on ImageNet-1k models are utilized.

\begin{figure*}[t!]
\vspace{-0.8cm}
\centering
\includegraphics[width=0.99\linewidth]{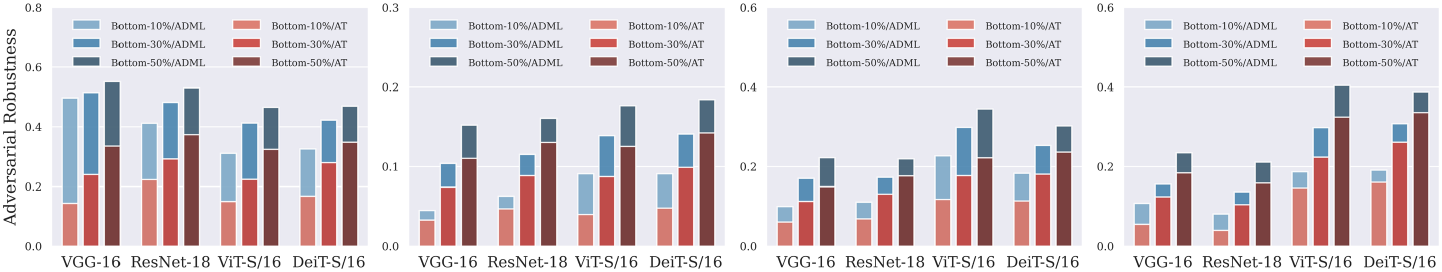}
\vspace*{-0.3cm}
\begin{flushleft}
    \hspace{1.3cm}{(a) CIFAR-10~\cite{krizhevsky2009learning} \hspace{1.5cm}(b) CIFAR-100~\cite{krizhevsky2009learning} \hspace{1.1cm}(c) Tiny-ImageNet~\cite{le2015tiny} \hspace{1.3cm}(d) ImageNet~\cite{deng2009imagenet}}
\end{flushleft}	
\vspace*{-0.15cm}
\caption{Cumulative distribution with averaged adversarial robustness for bottom-k classes against PGD~\cite{madry2018towards} on four benchmark datasets. Note that, $10\%$, $30\%$, and $50\%$ of $k$ values are applied, and perturbation budget is set to $[8/255, 8/255, 4/255, 2/255$] on each dataset.}
\vspace*{-0.2cm}
\label{fig:disparity}
\end{figure*}

\begin{figure}[t!]
\centering
\includegraphics[width=0.99\linewidth]{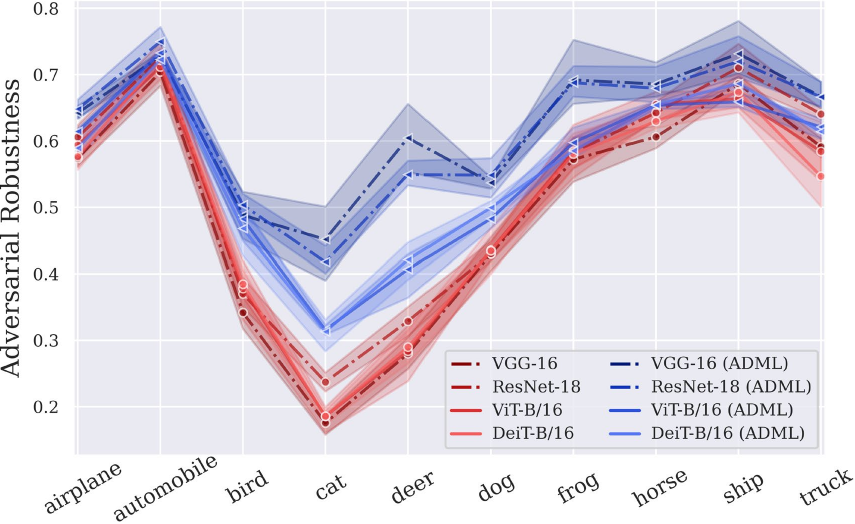}
\caption{Distribution of adversarial robustness across whole classes on CIFAR-10. Four methods: AT~\cite{madry2018towards}, TRADES~\cite{pmlr-v97-zhang19p}, MART~\cite{Wang2020Improving}, AWP~\cite{wu2020adversarial} are integrated on each architecture.}
\label{fig:cloud}
\end{figure}

\noindent\textbf{Training ADML.~} After the completion of standard adversarial training~\cite{madry2018towards}, we apply AT-based defense methods to line 4 in Algorithm~\ref{alg:1} for ADML. We then optimize adversarially trained CNNs in $10$ epochs using SGD~\cite{ruder2016overview} with a learning rate of $0.001$ scheduled by Cyclic~\cite{smith2017cyclical}, which allows empirically sufficient convergence to robustness. In addition, adversarially trained Transformers are also optimized with ADML using a learning rate of $0.0001$ on the equal experimental setup of CNNs. Note that, we set sample-splitting ratio in half (see Appendix G) for each batch, and cross-fitting is satisfied during training iterations.

\begin{table}[t!]
\centering
\renewcommand{\tabcolsep}{0.9mm}
\resizebox{1\linewidth}{!}{
\begin{tabular}{lccccccccccc}
\Xhline{3\arrayrulewidth}\rule{0pt}{9pt}
                           &  & PORT & {\footnotesize+ADML} & Gowal & {\footnotesize+ADML} & HAT  & {\footnotesize+ADML} & SCORE & {\footnotesize+ADML} & Wang   & {\footnotesize+ADML} \\
\midrule
\multirow{2}{*}{\rotatebox[origin=c]{90}{C-10}}  & Clean  & 87.0 & \textbf{88.1}  & 86.0  & \textbf{87.4}  & 88.2 & \textbf{89.5}  & 88.0  & \textbf{89.9}  & 91.4 & \textbf{91.8}  \\
\cdashline{2-12}\noalign{\vskip 0.5ex}
                           & AA     & 60.6 & \textbf{66.4}  & 60.7  & \textbf{66.8}  & 61.0 & \textbf{67.5}  & 61.1  & \textbf{68.0}  & 64.0 & \textbf{70.5}  \\
\midrule
\multirow{2}{*}{\rotatebox[origin=c]{90}{ C-100}} & Clean  & \textbf{65.9} & 65.8  & 59.2  & \textbf{59.9}  & 62.2 & \textbf{62.4}  & 62.0  & \textbf{62.3}  & 68.1 & \textbf{68.2}  \\
\cdashline{2-12}\noalign{\vskip 0.5ex}
                           & AA     & 31.2 & \textbf{37.9}  & 30.8  & \textbf{37.7}  & 31.2 & \textbf{37.3}  & 31.2  & \textbf{37.1}  & 35.7 & \textbf{41.1} \\
                           \Xhline{3\arrayrulewidth}\rule{0pt}{9pt}
\end{tabular}
}
\vspace{-0.38cm}
\caption{Comparing adversarial robustness of TRADES~\cite{pmlr-v97-zhang19p} using synthetic images: DDPM~\cite{ho2020denoising} and EDM~\cite{karras2022elucidating} whether to the inclusion of ADML for CIFAR-10/100 with WRN-34-10: PORT~\cite{sehwag2022robust} and WRN-28-10: Gowal~\cite{gowal2021improving}, HAT~\cite{rade2022reducing}, SCORE~\cite{pang2022robustness}, Wang~\cite{wang2023better}.}
\label{tab:aug}
\end{table}


\subsection{Robustness Validation on ADML}
\label{subsec:adv_robust}

\noindent\textbf{Adversarial Robustness.~} Based on our experimental setups, we have conducted enormous validations of adversarial robustness on CNNs in Table~\ref{table:cnn} and Transformers in Table~\ref{table:transformer}. As shown in these tables, employing ADML on AT-based defense methods: AT~\cite{madry2018towards}, TRADES~\cite{pmlr-v97-zhang19p}, MART~\cite{Wang2020Improving}, AWP~\cite{wu2020adversarial} enables to largely improve adversarial robustness, compared with that of each defense method baseline. Bai~\etal~\cite{bai2021transformers} have argued that Transformers cannot show noticeable adversarial robustness than CNNs, but we want to point out that the robustness of Transformers can be remarkably improved, especially in larger datasets. 

\noindent\textbf{Ablation Study.~} In Table~\ref{table:ablation}, we conduct ablation studies on the effect of sample-splitting plus cross-fitting on robustness and the effect of considering treatments as worst examples, non-worst examples, or both on robustness, either. According to the results, only considering treatments as worst examples can catch actual adversarial vulnerability, thereby improving robustness much more than others. 

\noindent\textbf{Utilizing Synthetic Images.~} Recently, several works~\cite{sehwag2022robust, gowal2021improving, rade2022reducing, pang2022robustness, wang2023better} have employed TRADES~\cite{pmlr-v97-zhang19p} utilizing synthetic images: DDPM~\cite{ho2020denoising} and EDM~\cite{karras2022elucidating} to improve adversarial robustness based on the insight that data augmentation such as CutMix~\cite{yun2019cutmix} can improve robustness~\cite{rebuffi2021data}. To further investigate the benefits of ADML, we experiment ADML combined with TRADES on the synthetic images. Table~\ref{tab:aug} shows ADML can further improve the robustness even on synthetic images, demonstrating its efficacy.

\subsection{Causal Analysis on ADML}
\noindent\textbf{Adversarial Vulnerability.~} To validate the alleviation of adversarial vulnerability existing in certain classes as in Figure~\ref{fig:intro}, we evaluate the averaged adversarial robustness for the cumulative distribution of bottom-$k$ classes with respect to the network prediction. We set the $k$ value as 10\%, 30\%, and 50\%. As in Figure~\ref{fig:disparity}, we can observe that AT shows noticeable vulnerability in bottom-$k$ classes, and such tendency pervades in four different datasets and architectures. If we successfully mitigate direct causal parameter of adversarial perturbations on each class, we expect apparent improvements of robustness for bottom-$k$ classes. As in the figure, we can observe the notable robustness of ADML in the vulnerable bottom-$k$ classes and corroborate its effectiveness to alleviate aforementioned phenomenon existing in current AT-based defenses. Further infographic is illustrated in Figure~\ref{fig:cloud} for the integrated distribution of baselines~\cite{madry2018towards, pmlr-v97-zhang19p, Wang2020Improving, wu2020adversarial} and their corresponding ADML adoptions on each architecture, and it shows further adversarial robustness in general (Additional results in Appendix H).


\begin{table}[t!]
\centering
\renewcommand{\tabcolsep}{2.0mm}
\resizebox{0.99\linewidth}{!}{
\begin{tabular}{lcccccccc}
\Xhline{3\arrayrulewidth}\rule{0pt}{9pt}
\multirow{2}{*}{Networks} & \multicolumn{4}{c}{CIFAR10}                    & \multicolumn{4}{c}{Tiny-ImageNet}              \\ \cmidrule(lr){2-5} \cmidrule(lr){6-9}
                       & $\rho_{10}$ & $\rho_{30}$ & $\rho_{50}$ & $\rho_\text{Avg}$ & $\rho_{10}$ & $\rho_{30}$ & $\rho_{50}$ & $\rho_\text{Avg}$ \\ 
\midrule
ResNet-18              & 53.98     & 56.89     & 52.97     & 67.33      & 4.93      & 5.39      & 5.54      & 6.49       \\
\cdashline{1-9}\noalign{\vskip 0.5ex}
WRN-28-10              & 63.60     & 68.70     & 68.27     & 77.86      & 6.72      & 5.91      & 6.60      & 10.74      \\
\cdashline{1-9}\noalign{\vskip 0.5ex}
ViT-B/16               & 76.78     & 85.45     & 80.03     & 84.33      & 1.54      & 1.73      & 1.91      & 3.08       \\
\cdashline{1-9}\noalign{\vskip 0.5ex}
DeiT-B/16              & 69.36     & 74.46     & 68.63     & 69.71      & 1.15      & 1.13      & 1.22      & 2.05 
\\
\Xhline{3\arrayrulewidth}\rule{0pt}{9pt}
\end{tabular}
}
\vspace{-0.2cm}
\caption{Relative ratio of causal parameter (\%) in CIFAR-10 and Tiny-ImageNet with four architectures. Note that $k$ is set to $10$, $30$, $50$, and Avg indicates average on whole classes in each dataset.}
\label{table:disparity}
\end{table}

\noindent\textbf{Causal Parameter.~} By deploying ADML, we present a way of mitigating the magnitude of causal parameter $|\theta|$. To numerically calculate $|\theta|$, we employ Eq.~(\ref{eqn:adml_theta}) and measure the average of $|\theta_\text{ADML}|$ for ADML with respect to the bottom-$k$ and whole classes, respectively. By dividing $|\theta_\text{ADML}|$ with $|\theta_\text{AT}|$, we can obtain relative ratio of causal parameter, $\rho_{k}:=100\times|\theta_\text{ADML}|/|\theta_\text{AT}|$ of adversarial examples in bottom-$k$ classes. This ratio indicates that relative intensity of causal parameter compared to that of AT~\cite{madry2018towards}. As in Table~\ref{table:disparity}, we can observe that ADML shows less intensity of the causal parameter than AT, which means less causal effects of adversarial perturbations on target classes. From combinatorial results of preceding robustness comparison in Sec.~\ref{subsec:adv_robust}, we corroborate that ADML indeed mitigate the intrinsic causal parameter and alleviate empirical observation in Figure~\ref{fig:intro}, thus results in adversarial robustness.
%

\section{Conclusion}
\label{sec:conclusion}
In this paper, we observe adversarial vulnerability varies across targets and still pervades even with deeper architectures and advanced defense methods. To fundamentally address it, we build causal perspective in adversarial examples and propose a way of estimating causal parameter representing the degree of adversarial vulnerability, namely Adversarial Double Machine Learning (ADML). By minimizing causal effects from the vulnerability, ADML can mitigate the empirical phenomenon as well as solidly improve adversarial robustness. Through intensive experiments, we corroborate the effectiveness of ADML for robust network.  

{\small
\bibliographystyle{ieee_fullname}
\bibliography{main}
}

\end{document}